\documentclass[10pt,twocolumn,letterpaper]{article}

\usepackage[pagenumbers]{cvpr} % To force page numbers, e.g. for an arXiv version

\makeatletter
\@namedef{ver@everyshi.sty}{}
\makeatother
\usepackage{tikz}

\usepackage[accsupp]{axessibility}  % Improves PDF readability for those with disabilities.

\usepackage{graphicx}
\usepackage{amsmath}
\usepackage{amssymb}
\usepackage{booktabs}
\usepackage{tabularx}
\usepackage{pgfplots}
\usepackage{environ}
\usepackage{multirow}
\usepackage{xcolor,pifont}
\usepackage{xspace}

\usepackage[pagebackref,breaklinks,colorlinks]{hyperref}
\usepackage{enumitem}

\usepgfplotslibrary{colorbrewer}
\newcommand{\greencmark}{\color{green}\checkmark}
\newcommand{\redxmark}{\color{red}\ding{55}}
\newcommand{\ourmodel}{\textsc{Gobbet}\xspace}
\newcommand{\dacc}{$\Delta \text{Acc}$\xspace}
\newcommand{\header}[1]{\noindent \textbf{#1}}

\usepackage[capitalize]{cleveref}
\crefname{section}{Sec.}{Secs.}
\Crefname{section}{Section}{Sections}
\Crefname{table}{Table}{Tables}
\crefname{table}{Tab.}{Tabs.}

\def\cvprPaperID{1} % *** Enter the CVPR Paper ID here
\def\confName{O-DRUM@CVPR}
\def\confYear{2022}

\pgfplotsset{compat=1.17}
\begin{document}

\makeatletter
\newsavebox{\measure@tikzpicture}
\NewEnviron{scaletikzpicturetowidth}[1]{%
  \def\tikz@width{#1}%
  \def\tikzscale{1}\begin{lrbox}{\measure@tikzpicture}%
  \BODY
  \end{lrbox}%
  \pgfmathparse{#1/\wd\measure@tikzpicture}%
  \edef\tikzscale{\pgfmathresult}%
  \BODY
}
\makeatother

%%%%%%%%% TITLE - PLEASE UPDATE
\title{Good, Better, Best: Textual Distractors Generation for \protect\\ Multiple-Choice Visual Question Answering via Reinforcement Learning}

\author{Jiaying Lu\textsuperscript{\ding{171}\ding{170}}\thanks{Work was done as a visiting researcher at Arizona State University.}, Xin Ye\textsuperscript{\ding{170}}, Yi Ren\textsuperscript{\ding{169}}, Yezhou Yang\textsuperscript{\ding{170}}\\
\textsuperscript{\ding{171}}Department of Computer Science, Emory University\\
\textsuperscript{\ding{170}}School of Computing and Augmented Intelligence, Arizona State University\\
\textsuperscript{\ding{169}}School for the Engineering of Matter, Transport and Energy, Arizona State University\\
{\tt\small jiaying.lu@emory.edu, \{xinye1, yiren, yz.yang\}@asu.edu}
% For a paper whose authors are all at the same institution,
% omit the following lines up until the closing ``}''.
% Additional authors and addresses can be added with ``\and'',
% just like the second author.
% To save space, use either the email address or home page, not both
}
% must include authlib package if using following

\maketitle
%%%%%%%%% ABSTRACT
\begin{abstract}
Multiple-choice VQA has drawn increasing attention from researchers and end-users recently. As the demand for automatically constructing large-scale multiple-choice VQA data grows, we introduce a novel task called \textit{textual Distractors Generation for VQA} (DG-VQA) focusing on generating challenging yet meaningful distractors given the context image, question, and correct answer. The DG-VQA task aims at generating distractors without ground-truth training samples since such resources are rarely available. To tackle the DG-VQA unsupervisedly, we propose \ourmodel, a reinforcement learning(RL) based framework that utilizes pre-trained VQA models as an alternative knowledge base to guide the distractor generation process. In \ourmodel, a pre-trained VQA model serves as the environment in RL setting to provide feedback for the input multi-modal query, while a neural distractor generator serves as the agent to take actions accordingly. We propose to use existing VQA models' performance degradation as indicators of the quality of generated distractors. 
On the other hand, we show the utility of generated distractors through data augmentation experiments, since robustness is more and more important when AI models apply to unpredictable open-domain scenarios or security-sensitive applications. We further conduct a manual case study on the factors why distractors generated by \ourmodel can fool existing models.
\end{abstract}

%%%%%%%%% BODY TEXT
\section{Introduction}
\label{sec:intro}

\begin{figure}[htb]
\centering
\includegraphics[width=0.8\linewidth]{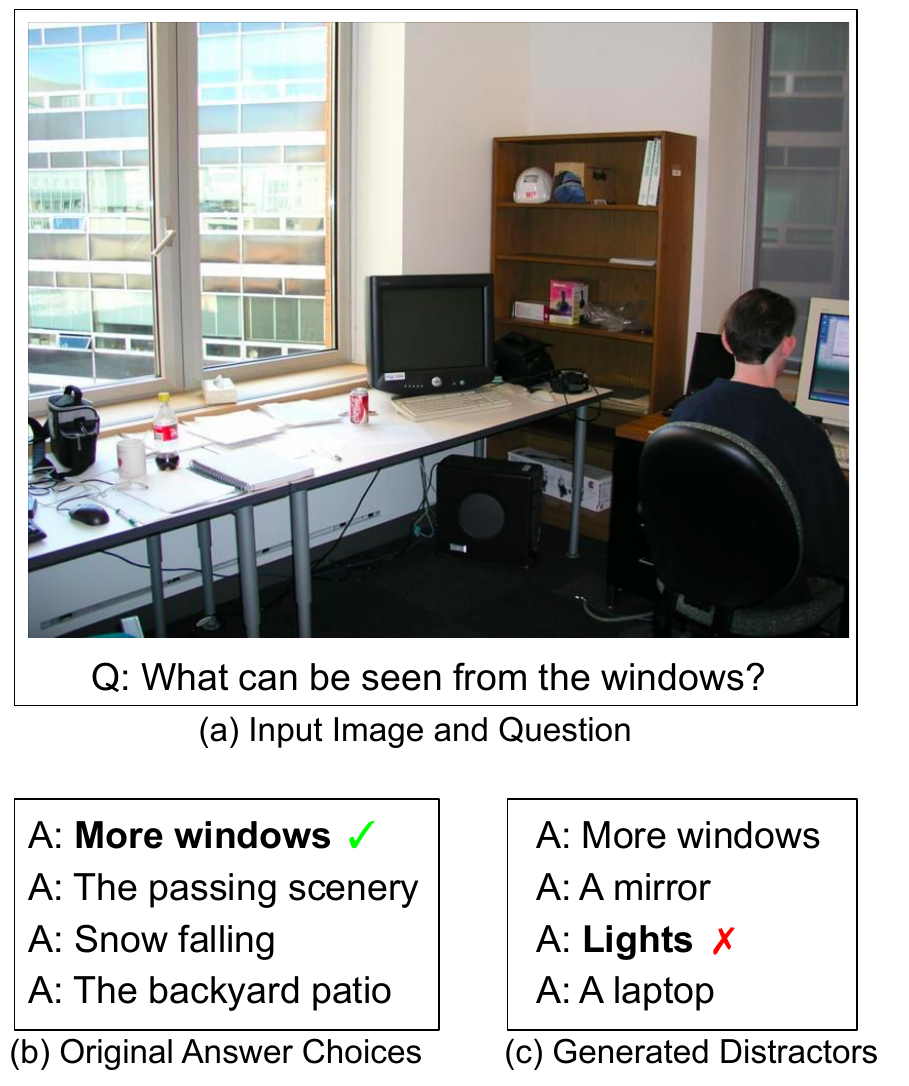}
\vspace{-0.25cm}
\caption{An example of DG-VQA task. The well-trained VQA model predicts the right answer choice for the input image and question (\ref{fig:example_DGVQA}a and \ref{fig:example_DGVQA}b). However, it is easy to distinguish correct choice from distractor choices. The model will be fooled when encountering generated distractors (\ref{fig:example_DGVQA}c). }
\label{fig:example_DGVQA}
\end{figure}

Visual Question Answering (VQA)~\cite{AntolVQA,zhu2016visual7w,okvqa,zellers2018recognition} is an emerging research problem that requires algorithms to answer arbitrary natural language questions about a given image. Recently, VQA has attracted a large number of interests across computer vision, natural language processing, and knowledge representation and reasoning communities, since these questions require AI models' capability of understanding vision, language, and even external knowledge~\cite{okvqa,zellers2018recognition} to answer. In general, VQA can be divided into two specific sub-tasks according to the question forms: 1) open-ended VQA~\cite{AntolVQA,okvqa} requiring a free-form response; and 2) multiple-choice(MC) VQA~\cite{zhu2016visual7w,zellers2018recognition} requiring a single answer picking from a list of given candidates. In this paper, we are particularly interested in the MC VQA task.

With the advancement of deep learning, neural network models have achieved remarkable progress to bridge the gap between human performance and state-of-the-art neural network models on the MC VQA task. 
However, it has been pointed out that the distractors (the wrong candidate choices) are too simple or biased~\cite{jabri2016revisiting} in several benchmark datasets, which raises doubt about the proposed models' actual discriminative ability. For instance, in the toy example shown in Figure~\ref{fig:example_DGVQA}, the correct answer is ``More windows'', while the distractor ``The passing scenery'' is not challenging because the scenery outside is stationary. Similarly, ``Snow falling'' is not challenging due to the sunny weather.

To facilitate multiple-choice VQA better to serve as a robust ``multi-modality Turing test''~\cite{turing2009computing,geman2015visual} and to explore factors causing failure of existing VQA models, we introduce a novel task as generating challenging distractors, dubbed as DG-VQA: \textit{textual Distractor Generation for VQA}. The task can be defined as follows: Given an image with a corresponding natural language question and a correct answer, generating distractors that lead to trained VQA models failing at picking the right choice from the candidate list, where the candidate list is composed of the correct answer and generated distractors. 
Figure~\ref{fig:example_DGVQA}c provides a toy example of generated distractors, where ``A mirror'' and ``Lights'' are very confusing choices for both AI models and humans. Producing such distractors provides a tool for researchers to figure out whether well-trained VQA models are vulnerable to potential attacks and determine whether they are ready for real-world deployment. Moreover, MC question answering is widely used in the education area, and manual distractor generation is hard and time-consuming. There are some previous works \cite{liang2018distractor,gao2018generating} focusing on automatic distractor generation (DG) to alleviate instructors' workload. Unfortunately, none of them consider that applying multimodal materials in education becomes increasingly favorite.

One of the major technical challenges for the proposed DG-VQA task is that the training data is very limited or not available in real-world scenarios. Previous unsupervised distractor generation methods typically rely on the similarity measurements between the textual answer and generated candidates, thus ignoring the critical signals from the image input. Owing to the recent progress of pre-trained deep neural networks on large datasets, the pre-trained VQA models are capable of generating high-quality answers according to the given image and question. Whilst learning the correct answers, these models may also be storing plausible responses to the multimodal context, and may be able to generate distractors for the input. Therefore, we propose to utilize these existing VQA models as an alternative knowledge source to guide a distractor generator, which helps train the distractor generator without training samples. More specifically, we fix the pre-trained VQA models and use them to produce numerical quality judgment for the generated distractors based on input context (\textit{i.e.} the judgment score represents which distractor is better). To propagate the non-differentiable judgment scores to the distractor generator, we opt for the reinforcement learning (RL) techniques \cite{rennie2017self,li2018learning}. We dub the proposed framework as \ourmodel (``GOod Better BEsT'' for DG-VQA). In \ourmodel, the distractor generator, which is regarded as an agent, receives rewards from the pre-training VQA model, which serves as the environment, based on the input context and generated distractors, which are actions taken by the agent. Therefore, the distractor generator is trained to maximize the cumulative reward from the pre-training VQA model. The choice of reward function is flexible as long as it represents the quality of generated distractors. In practice, we define the negative judgment score as the reward, and we utilize the policy gradient algorithm to optimize the generator.

In this work, an extensive suite of experiments has been conducted on the public MC VQA benchmark Visual7W~\cite{zhu2016visual7w}. Since the goal is to generate challenging distractors that lure existing models to fail, we propose to adopt performance degradation as the main measurement for generated distractors. Through experiments results on different existing VQA models, we validate distractors predicted by \ourmodel outperform all baseline methods. In addition, we further demonstrate the utility of generated distractors by feeding them as augmented data into VQA models. We have observed the performance boosts on models trained with augmented data, which support the effectiveness of \ourmodel from another perspective. Finally, we conduct case studies on distractors created by baselines and \ourmodel, to gain a more intuitive sense of why \ourmodel can generate more challenging distractors than other methods.

% \jiaying{Contribution of DG-VQA}
% The contribution of this paper is three-fold:

% \begin{enumerate}
%     \item We introduce a new DG-VQA task for vision and language understanding research, accompanied with a practical metric for evaluating the quality of the generated distractors. The DG-VQA task also provides a novel perspective for the interpretability of the exisiting neural models' decision boundary. 
%     \item We propose a novel perspective to formulate DG-VQA task as a reinforcement learning task and optimize it with policy gradient. The proposed RL model addresses the lack of training data issue.
%     \item We present and show that by incorporating the challenging distractors one can train a more robust VQA model.
% \end{enumerate}
\section{Related Work}

\noindent \textbf{Visual Question Answering.} The open-ended answering task~\cite{johnson2017clevr,sampat2021clevr_hyp} and the multiple-choice task~\cite{zhu2016visual7w,zellers2018recognition} are two typical tasks for VQA~\cite{AntolVQA}. In this work, we focus on the multiple-choice task. Existing VQA models commonly combine an image encoder and a textual encoder to represent input pictures and input questions. The multi-modal context embedding is fused and then fed into an answer decoder to generate the answers. Traditionally, convolutional neural networks~\cite{he2016deep} and recurrent neural networks~\cite{hochreiter1997long} are popular choices for image encoders and textual encoders, while the answer decoder ranges from a softmax classifier~\cite{fukui2016multimodal}, an RNN decoder~\cite{malinowski2015ask} to a dot product layer~\cite{jabri2016revisiting}. More recently, Transformer-based networks~\cite{li2019visualbert} have shown distinguish performance as a uniform layer of both multi-modality encoders and decoders.

\noindent \textbf{Distractor Generation.} Automatic distractors generation (DG) from text is explored in-depth in the Natural Language Processing domain. At the same time, there are only a few studies in the multi-modal domain. Most prior approaches to textual DG are based on unsupervised similarity measures. These include n-gram co-occurrence likelihood~\cite{hill2016automatic}, word/sentence embedding-based semantic similarities~\cite{kumar2015revup}, syntactic homogeneity~\cite{chen2006fast} and ontology-based similarity~\cite{stasaski2017multiple}. Besides, other works utilize supervised learning algorithms for DG. Sakaguchi \textit{et al.} ~\cite{sakaguchi2013discriminative} train a discriminative model to predict distractors, Liang \textit{et al.}~\cite{liang2018distractor} apply learning to rank algorithm, and Gao \textit{et al.}~\cite{gao2018generating} use an end-to-end framework to produce distractors generatively. Although being successful, multimodality knowledge is still required to produce high-quality distractors.

% \noindent \textbf{Adversarial Examples.} Szegedy et al.~\cite{szegedy2013intriguing} present an initial investigation on the existence of adversarial examples. Many authors then discuss adversarial examples for image classification~\cite{goodfellow2014generative,nguyen2015deep}. Recent work has demonstrated that adversarial examples widely exist in various fields, including text classification~\cite{wong2017dancin,alzantot2018generating}, autonomous cars~\cite{sitawarin2018darts}, and so on. For image, typical approaches are gradient-based methods such as fast gradient sign method (FGSM)~\cite{goodfellow2014generative} and its variants. However, algorithms used for adversarial image samples generation can not easily transfer to the text domain. The discrete nature of natural language makes small perturbations easily perceptible, and sometimes the replacement of one word would change the semantic of the sentence dramatically. Adversarial text example requires the perturbated text to cause misclassification while preserving semantics. The goal of DG-VQA task is similar but at the opposite end. Instead of preserving semantics, distractors should mislead models towards misclassification with different semantic meaning. %Otherwise, distractors would become another ``correct'' alternative. 

\noindent \textbf{Pre-trained Models as Knowledge Bases.}Knowledge bases have shown great potential in multi-modal information retrieval setting~\cite{okvqa,zhu2015building,luo2021weakly}. Unfortunately, the construction of a large-scale multi-modal knowledge base (KB) is time-consuming, and the coverage of KB is limited. Hence, in practice, we often need to populate these KBs from raw text or other modalities, where ad-hoc complex pipelines are required and noise can easily accumulate. Recently, researchers start to explore alternative light-weight KBs. Gokhale \textit{et al.}~\cite{gokhale2021semantically} incorporate the semantics-inverting and semantics-preserving transformations over input textual query for the robust vision-and-language model optimization, instead of explicit knowledge of the text. Petroni \textit{et al.}~\cite{petroni2019language} proposes to utilize language models pre-trained on large textual corpora as KBs storing relational linguistic knowledge, and experiments on multiple downstream tasks such as question answering and relation prediction well support their arguments. Furthermore, Wang \textit{et al.}~\cite{wang2020language} explore constructing open knowledge base from pre-trained language models without human supervision. Our \ourmodel share a similar idea to these works to use pre-trained VQA models as alternative KBs to retrieve distractors based on input multi-modal query.

\noindent \textbf{Reinforcement Learning.} Reinforcement learning (RL)~\cite{sutton2018reinforcement} has been adopted in a variety of vision and language tasks, such as image captioning~\cite{rennie2017self}, text to image synthesis~\cite{reed2016generative}, VQA~\cite{liu2018ivqa,fan2018reinforcement} and visual dialogue~\cite{zhang2017asking}. Liu et al.~\cite{liu2018ivqa} propose a RL-based strategy to generate visual questions. Fan et al.~\cite{fan2018reinforcement} enhance content and linguistic attributes of produced questions by introducing two discriminators in an RL framework. In \ourmodel, we utilize the REINFORCE algorithm~\cite{williams1992simple} to propagate the feedback backward from the pre-trained VQA models to the distractor generator.
\section{Problem Definition}
\label{sec:problem_def}

Textual Distractor Generation for multiple-choice VQA (DG-VQA) aims at generating challenging distractors (wrong options) $\mathbf{D}=\{\mathbf{d}_1, \mathbf{d}_2, ..., \mathbf{d}_k\}$ based on the input image $\mathbf{i}$, question $\mathbf{q}$ and answer $\mathbf{a}$.  Both $\mathbf{q}$, $\mathbf{a}$ and $\mathbf{d}$ are all textual sequence consist of words $w_{1:T}=(w_1, w_2, ..., w_T)$.  Depending on the dataset, sometimes not only the correct answer $\mathbf{a}^{\text{COR}}$ is provided, but also several wrong answers $\{\mathbf{a}^{\text{WOR}}_1, ..., \mathbf{a}^{\text{WOR}}_m\}$. In this work, we focus on the most general case that only one correct answer $\mathbf{a}$ is provided. Figure~\ref{fig:example_DGVQA} displays a toy example of proposed DG-VQA task. %where the input image $\mathbf{i}$ depicts somebody's office with a window, the input question is ``What can be seen from the windows?'', and the input correct answer $\mathbf{a}$ is ``More windows''. 
The generated distractors are expected to be challenging, since such distractors can better serve as an effective assessment for humans and AIs~\cite{goodrich1977distractor,jabri2016revisiting}. However, challenging does not mean the generated distractors $\mathbf{D}$ must be semantically equivalent to the input correct answer $\mathbf{a}$. Therefore, we propose pre-trained models' performance degradation and data augmentation improvement as two indicators to evaluate the generated distractors.
%We will present details about these two measurements in $\S$\ref{sec:results}.
\section{\ourmodel: A Reinforcement Learning Framework for DG-VQA}
\label{sec:approach}

\begin{figure}[htbp!]
\centering
\includegraphics[width=\columnwidth]{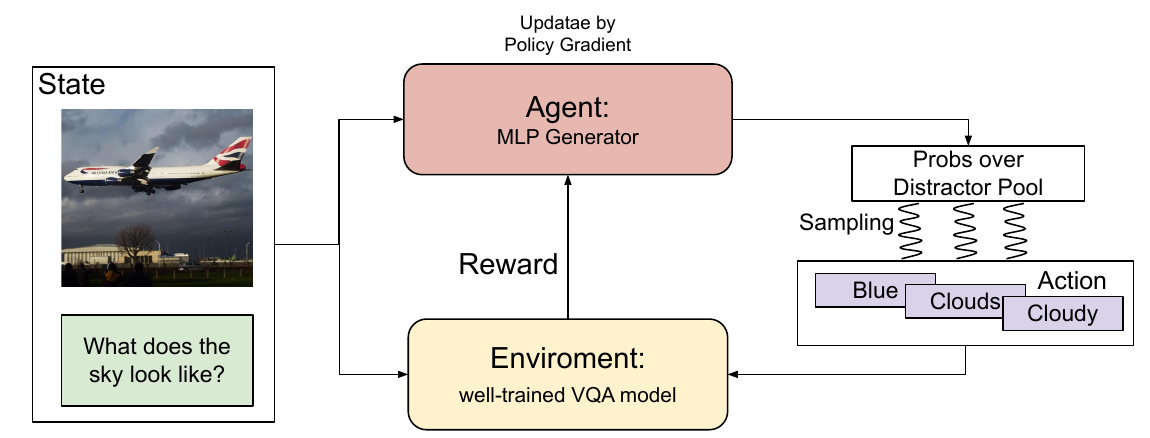}
\caption{The proposed \ourmodel Framework}
\label{fig:framework}
\end{figure}

To tackle the DG-VQA task without direct training samples, our key insight is to leverage the pre-trained VQA models as an alternative knowledge source. Therefore, we propose \ourmodel (``GOod Better BEsT'' for DG-VQA), a reinforcement learning-based framework where the agent model is trained to generate distractors based on the feedback from the environment. Figure~\ref{fig:example_DGVQA} shows the overall architecture of the proposed \ourmodel. We will introduce the technical details in the following subsections.

%Figure \ref{fig:example_DGVQA} displays an example of the DG-VQA task. Formally, given an image $i$, a natural language question $q$ and four corresponding multi-choice answers $a$s, which include one correct answer $a^{correct}$ and three wrong answers $a^{wrong}$s in original dataset, the task is to produce three plausible but incorrect distractors $d$s. A DG-VQA is learned so that its generated distractors maximize the expected accuracy drop on a given multi-choices VQA model, where accuracy is measured as the percentage of times the model picks up the correct choice. 

\subsection{DG-VQA as A RL Problem}

Inspired by recent progress in reinforcement learning (RL) and adversarial generation~\cite{moosavi2016deepfool,yao2019improving}, RL methods are promising for scarce supervision scenarios and efficient to address the inconsistency between the training objective and test metrics~\cite{fan2018reinforcement,zhang2017asking}. Therefore, we adopt a policy gradient framework \ourmodel to generate textual distractors for multiple-choice VQA. \ourmodel has two major components: (1) the agent $G_{\theta}$, which is a distractor generator that generates high-quality distractors $\mathbf{D}$ according to the input image $\mathbf{i}$ and question $\mathbf{q}$; (2) the environment $J_{\phi}$ which is a pre-trained VQA model that produces rewards based on the generated distractors and the input context. \ourmodel is somehow similar to the GAN framework~\cite{goodfellow2014generative} if we regard the agent as the generator and the environment as the discriminator, but we opt to fix the pre-trained VQA model to serve as the static external knowledge source during the \ourmodel training process. The reason that we do not make the VQA models trainable is the concern of local convergence \cite{Mescheder2018ICML}.

%Here we utilize a well-trained model as the discriminator rather than train a model from scratch. The reason behind it is the concern of local convergence \cite{Mescheder2018ICML}. We put our approach under a semi white-box attack setting where $G_{\theta}$ can receive feedback signals regarding selected choices from $J_{\phi}$, but can not access $J_{\phi}$ parameters or gradients. From the RL perspective, the well-trained VQA model $J_{\phi}$ serves as the environment, and the generative model $G_{\theta}$ is the policy agent.

We first denote distractors generation as a sequence generation process. The distractor generator $G_{\theta}$ is trained to produce a set of distractors $\mathbf{D}=\{\mathbf{d}_1,\mathbf{d}_2,...,\mathbf{d}_k\}$, where each $\mathbf{d}$ is a seuqnce of words $\mathbf{d}=d_{1:T}=(d_1,d_2,...,d_t,...,d_T)$. It is worth noting that, the bold math symbol $\mathbf{d}_k$ denotes the $k-th$ distractor in a set of distractors $\mathbf{D}$, while the regular math symbol $d_t$ denotes the $t$-th word in a sequence of word. At each timestep $t$, $G_{\theta}$ is generate one word $d_t$ given the input image $\mathbf{i}$, the question $\mathbf{q}$, and the generated distractor sequence untial last timestep $d_{1:t-1}$:
\begin{equation}
  d_t = G_{\theta}(\mathbf{i}, \mathbf{q}, d_{1:t-1}).
\end{equation}
%Since $G_{\theta}$ outputs a probability distribution over each token in produced sequence, we can use decoding algorithms like greedy search or beam search to locate the top-3 distractors. 
\noindent Under the RL setting, at timestep $t$ the state $s$ of the generator is the currently produced tokens $d_{1:t-1}$ and the action $a$ is the next token $d_t$ to produce. So the state transition is deterministic once an action has been chosen. Following the notation in \cite{sutton2018reinforcement}, the object of the $G_{\theta}$ is to produce a sequence to minimize its negative expected reward:
\begin{equation}
    L(\theta)=-\mathbb{E}_{d_{1:T}\sim G_{\theta}}[R(d_{1:T})],
\end{equation}
\noindent where $d_{1:T}$ is the a sampled generation from the model $G_{\theta}$.

Without the loss of generality, we adopt the REINFORCE algorithm~\cite{williams1992simple} to optimize the agent $G_{\theta}$ through the policy gradient, and we take judgement scores $R(d_{1:T})$ (\textit{i.e.} the likelihood of $\mathbf{d}$ as the answer to the input context $(\mathbf{i}, \mathbf{q})$) from the environment $J_{\phi}$. Formally, the optimization process is defined as follows:
\begin{equation}
\begin{aligned}
    d_{1:T} &= G_{\theta}(\mathbf{i},\mathbf{q}; d_{1:T-1}), \\ 
    R(d_{1:T}) &= J_{\phi}(\mathbf{i}, \mathbf{q}, d_{1:T}), \\
    \nabla_{\theta}L(\theta) &= -\mathbb{E}_{d_{1:T}\sim G_{\theta}}[R(d_{1:T})\nabla_{\theta}log G_{\theta}(d_{1:T})].
\end{aligned}
\label{eqn:rl_eqs}
\end{equation}
\noindent It is worth mentioning that the environment $J_{\phi}$ can only output a reward value from a completed sequence $\mathbf{d}=d_{1:T}$. However, in DG-VQA setting and under the sequence generation scenario, the model should consider the partial reward of the incompleted sequence $R(d_{1:t}), \forall t<T$. To tackle this challenge, we follow the common practice to use the Monte Carlo search~\cite{williams1992simple} to sample the unknown last $T-t$ tokens at intermediate timesteps. In practice, the expected gradient can be approximated using several distractors $\mathbf{d}^s$ sampled from $G_{\theta}$ for each input image, question, and correct answer triplet in a minibatch.
\begin{equation}
\label{eqn:gradient_practical}
    \nabla_{\theta}L(\theta) \approx \sum_{s} -R(\mathbf{d}^s)\nabla_{\theta}log G_{\theta}(\mathbf{d}^s).
\end{equation}

\subsection{The Agent: A neural distractor generator}

\begin{figure}[htbp!]
\centering
\includegraphics[width=0.9\linewidth]{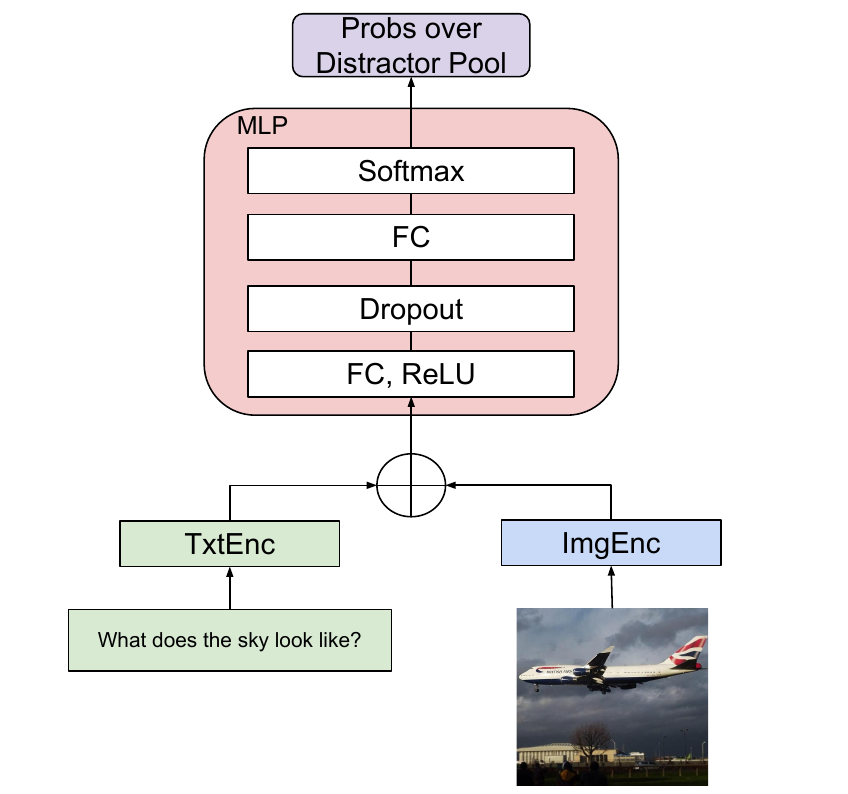}
\caption{The model architecture for agent in \ourmodel, where TxtEnc denotes text encoder, ImgEnc denotes image encoder, and FC denotes fully connected layer.}
\label{fig:mlp_architecture}
\end{figure}

In \ourmodel framework, the agent is responsible for generating textual distractors according to the rewards from the environment. Therefore, the technical choice of the agent is flexible, as long as it can select actions based on observations from the environment and can update its policy parameters. From the possible choices such as MLP(multi-Layer Perceptron), RNN, and Transformer, we select MLP for its advantages in training speed. Moreover, empirical results show that the simple MLP model is sufficient to generate challenging distractors under the guidance of the environment. Consequently, one complete sequence $d_{1:T}$ can be generated for each training iteration by selecting the output distractor over a distractor candidate pool. Thus, the distractor generator can be formulated as follows:
\begin{equation}
    \mathbf{d} = d_{1:T} = G_{\theta}(\mathbf{i}, \mathbf{q}).
\end{equation}

Figure \ref{fig:mlp_architecture} depicts the details of the agent model's architecture, which is essentially a encoder-decoder model. The encoder is a two-channel encoder that embeds the input question and input image into dense embedding vectors, while the decoder is a MLP that outputs the probabilities over the candidate distractor pool. For the encoder part, a text encoder and an image encoder are applied to the input question $\mathbf{q}$ and image $\mathbf{i}$, respectively. 
\begin{equation}
    \mathbf{x_q} = \mathcal{F}_{t}(\mathbf{q}),
    \mathbf{x_i} = \mathcal{F}_{i}(\mathbf{i}),
\end{equation}
\noindent where the text encoder $\mathcal{F}_{t}$ is responsible to produce the dense embedding of a textual sequence $\mathbf{q}=q_{1:T}$, thus MLP, CNN, RNN, or Transformers can be chosen. We implement the average of the pre-trained word embeddings~\cite{mikolov2013distributed} as the representation of $q$. Similarly, the technical choice of image encoder $\mathcal{F}_{i}$ is also flexible, and we opt to pre-trained deep CNN model~\cite{he2016deep}. 

% The architecture of our generative model $G_{\theta}$ is a multiple-layer perceptron. It is widely used in many domain and has shown success in visual question answering by Jabri's work \cite{jabri2016revisiting}. Figure \ref{fig:mlp_architecture} depicts the layers of the MLP model. Specifically, the questions are represented by averaging word2vec~\cite{mikolov2013distributed} embedding over all tokens. The images are represented using features computed by a pre-trained deep CNN image encoder. Unless otherwise stated, we use the penultimate layer of Resnet-101~\cite{he2016deep} as the extracted features. Then, the multi-modality feature sets are concatenated and used to train a classification model that predicts the corresponding distractor label. 

After we get the embedding $\mathbf{x_{q}}$ for input question and $\mathbf{x_i}$ for input image, we fuse them together as the overall input context representation by $\mathbf{c} = \mathbf{x_i} \oplus \mathbf{x_q}$, as shown in Figure~\ref{fig:mlp_architecture}. The fuse operation $\oplus$ can be concatenation, element-wise summation, element-wise multiplication or bilinear pooling~\cite{fukui2016multimodal}, where we choose concatenation in \ourmodel implementation. The context embedding $\mathbf{c}$ is then feed into the decoder, and we adopt a multilayer perceptron (MLP) for it: 
\begin{equation}
    \mathbf{z} = (w_2\ ReLu(w_1\mathbf{c} + b_1) + b_2),
\end{equation}
where $w_1,w_2,b_1,b_2$ denote the learnable parameters for the first or second layer in the MLP, $\mathbf{z}$ denotes the predicted unnormalized distribution over the distractor pool. Finally, the output probability over the distractor pool is obtained by $P(\mathbf{d}|\mathbf{i}, \mathbf{q}) = \text{softmax}(\mathbf{z})$. 

% A multilayer perceptron (MLP) is adopted as the classification model trained on the concatenated features: 1) The word2vec embedding (300-dimensional), and 2) the image features (2048-dimensional). By default, the MLP has 4,096 hidden units unless otherwise specified. We denote the image and question features as $x_i$ and $x_q$, respectively. By denoting the concatenation as $c = x_i \oplus x_q$, we formulate the models as follows:
% \begin{equation}
%     z = (w_2\ ReLu(w_1c + b_1) + b_2).
% \end{equation}
% The MLP outputs a distribution over the distractor pool using the softmax function. Then, the system selects the distractors $ds$ with the top-3 highest likelihood from $G_{\theta}(i, q)$.
% \begin{equation}\label{d_probabiltiy}
%     P(d|i, q) = softmax(z).
% \end{equation}

\subsection{The Environment: VQA Models}
\label{subsec:environment}

In \ourmodel framework, the environment $J_{\Phi}$ is responsible for providing rewards according to the input context and the actions made by the agent (\textit{i.e.} the distractors generated by the agent), as Equation~\ref{eqn:rl_eqs} defines. More specifically, $J_{\Phi}$ is a pre-trained multiple-choice VQA model that is fixed during the \ourmodel training and testing process. In principle, we take the validity score $J_{\Phi}(\mathbf{i}, \mathbf{q}, \mathbf{d})$ as the reward, as the validity score indicate to what extend the generated distractor $\mathbf{d}$ is a ``plausible'' answer to the input context. Furthermore, we punish the distractor $d$ which is semantically equivalent to the correct answer $\mathbf{a}$ for the given context. The semantic similarity module is a BERT~\cite{devlin2018bert} model trained on sentence similarity tasks, and it produces a binary label that represents whether the two input sentences are semantically similar. Therefore, we define the reward function as follows:

\begin{equation}
\label{eqn:reward}
    R(d) = \left\{
      \begin{array}{lr}
      -1 & if\ \text{IsSemEquiv}(\mathbf{d}, \mathbf{a});
      \\ J_{\phi}(\mathbf{i}, \mathbf{q}, \mathbf{d}) & \text{otherwise}.
      \end{array}
    \right.
\end{equation}

Any multi-choice VQA model which produces a validity (sometimes also called likelihood) scores of responses for given visual questions can serve as the environment in \ourmodel framework, such as TellingVQA~\cite{zhu2016visual7w}, RevisitedVQA~\cite{jabri2016revisiting}, MCB~\cite{fukui2016multimodal}, \textit{etc}.
It is worth noting that the choice of environment is not restricted to the abovementioned models, but is generally applicable to any VQA models which can produce such scores. Moreover, Our proposed \ourmodel also supports leveraging a bundle of pre-trained VQA models together as the environment to provide a combined reward. 
\section{Experiments}
\label{sec:results}

In this section, we evaluate our proposed \ourmodel model focusing on the following research questions:

\begin{itemize}
    \item \textit{RQ1}: How does \ourmodel perform in comparison to other methods?
    %in terms of fooling pre-trained VQA models?
    \item \textit{RQ2}: Can generated distractors help build more robust VQA models?
    \item \textit{RQ3}: How is the quality of generated distractors?
\end{itemize}

\subsection{Experimental Settings}
\header{Dataset.} We evaluate our model on \textit{Visual7w}~\cite{zhu2016visual7w}, which is a public multiple-choice visual question answering dataset. \textit{Visual7w} consists of 47,300 images from COCO and 327,939 multiple-choice QA pairs collected on Amazon Mechanical Turk.

\header{Evaluation Metrics.} Traditional metrics of distractor generation for question answering~\cite{pho2014multiple,liang2017distractorGAN,liang2018distractor}, such as reliability and validity, often rely on manual evaluation, which are hard to scale. In order to enable the automatic evaluation, we define the ability of generated distractors to fool pre-trained VQA models as the metric, denoted as \dacc. \dacc is the difference between VQA model's performance on the original distractors and on the generated distractors $\text{Acc}_{\text{original}} - \text{Acc}_{\text{generated}}$, \textit{i.e.} the performance degradation of VQA model when presenting generated distractors instead of original distractors. The higher \dacc is, the better-generated distractors are.  In this work, we leverage the following popular VQA models for calculating \dacc: 

\begin{itemize}
    \item \textit{\textbf{TellingVQA}}~\cite{zhu2016visual7w} is a recurrent QA model with spatial attention. It first encodes the image through a pre-trained VGG-16 model \cite{Simonyan15}. Then it uses a one-layer LSTM to read the image encoding and all the question tokens. It continues to feed the answer choice tokens into LSTM, and would finally produce the validity score.
    \item \textit{\textbf{RevisitedVQA}}~\cite{jabri2016revisiting} proposes a light architecture for MC VQA task. RevisitedVQA receives an image-question-answer triplet, encodes it, and utilizes a MLP to compute whether or not the triplet is correct.
    \item \textit{\textbf{MCB}}~\cite{fukui2016multimodal} proposes a novel method called Multimodel Compact Bilinear pooling to efficiently and expressively combine language and vision features. 
\end{itemize}

\header{Baseline Methods.} We compare our \ourmodel with the following baseline methods:

\begin{itemize}
    \item \textit{\textbf{Q-type prior}} is a heuristic method for distractor generation. We select three most popular answers per question type as distractors. 
    \item \textit{\textbf{Adversarial Matching}}~\cite{zellers2018recognition} forces distractors to be as relevant as possible to the context (image and question), while preventing distractors to be overly similar to the correct answer. We also employ BERT~\cite{devlin2018bert} to compute the relevance between the context and the distractor, and ESIM+ELMo~\cite{chen2017esim,ilic2018elmo} to compute the similarity between the answer and the distractor. During training, distractors are responses randomly sampled from the whole training response(answer choice) pool.
    \item \textit{\textbf{LSTM Q+I}}~\cite{AntolVQA} utilizes a two-layer LSTM to encode the input question and a VGGNet~\cite{Simonyan15} to encode the input image. After a point-wise multiplication operation to fuse question embedding and image embedding, a multi-layer perceptron is employed to predict the response over a pre-defined response pool. Similar to \textit{Adversarial Matching}, we construct the pool using all correct answers in the training set. We change the training targets from correct answers to incorrect ones. The incorrect responses are generated by pre-trained VQA models but discard the generated responses that are identical to correct answers.
\end{itemize}

\subsection{Implementation Details} 

For the agent in \ourmodel, we adopt a two-channel vision and language neural network that outputs probabilities over the candidate distractor pool. We set the candidate distractor frequency threshold to $20$, to filter the candidate pool size $K$ to $1516$, which covers $2\%$ of all training and validation choices. The questions are represented by 300-dim averaged word embeddings from the pre-trained fastText~\cite{bojanowski2017enriching} model. We use all words in the training dataset to finetune the word embedding. In the experiment, we set the dropout rate to $0.5$ in each hidden layer with a ReLU activation. We further set the maximum training epochs as $200$ with the early stop strategy.

For the environment in \ourmodel, we adopt RevisitedVQA model ~\cite{jabri2016revisiting} for its superior efficiency and effectiveness.  The pre-trained RevisitedVQA model outperforms other state-of-the-art models which are mentioned in $\S$~\ref{subsec:environment}, as it achieves $65.8\%$ accuracy on the Visual7W dataset. We evaluate the proposed \ourmodel with two ablated versions:

\begin{itemize}
    \item \textbf{\ourmodel-base}: Model parameters are updated only through policy gradient, where the rewards are from the pre-trained VQA models as the environment.
    \item \textbf{\ourmodel-warmup}: Reinforce algorithm is known to have a large variance. Inspired by Imitation Learning~\cite{ho2016generative} and Teacher Forcing~\cite{lecun2015deep}, we first train the agent model with correct answer choice using cross-entropy loss for a small size (80) epochs. The warmup training process is to prevent generating unstable results. Then we train the agent as in \ourmodel-base.
\end{itemize}

\begin{table*}[htbp!]
\small \centering
\caption{Pre-trained models performance on generated distractor for Visual7W dataset. The columns ``TellingVQA'', ``RevisitedVQA'', and ``MCB'' represent the VQA models that take the multiple-choice VQA assessment. The first row ``Original distractors'' indicates VQA models are presented to the original distractors, while other rows indicate the distractors are generated by corresponding DG-VQA models. \dacc denotes the performance degradation of pre-trained VQA models, and higher \dacc means the better generated distractors.}
\begin{tabular}{l|cc|cc|cc}
\hline
\multirow{2}{5em}{Model} & \multicolumn{2}{c|}{TellingVQA~\cite{zhu2016visual7w}} & \multicolumn{2}{c|}{RevisitedVQA~\cite{jabri2016revisiting}} & \multicolumn{2}{c}{MCB~\cite{fukui2016multimodal}} \\
& $Acc$ & $\Delta Acc$ & $Acc$ & $\Delta Acc$ & $Acc$ & $\Delta Acc$  \\
\hline \hline
Original distractors & 55.6\% & - & 64.8\% & - & 62.2\% & - \\
\hline
\multicolumn{7}{c}{Baselines} \\
\hline
\textit{Q-type prior} & 57.3\% & -1.7\% & 68.7\% & -3.9\% & 85.7\% & -23.5\% \\
\textit{Adversarial Matching}~\cite{zellers2018recognition} & 54.7\% & 0.9\% & 71.7\% & -6.9\% & 51.3\% & 10.9\% \\
\textit{LSTM Q+I}~\cite{AntolVQA} & 41.7\% & 13.9\% & 68.9\% & -4.1\% & 85.7\% & -23.5\% \\
\hline
\multicolumn{7}{c}{Proposed Methods} \\
\hline
Reward from RevisitedVQA \\
  - \ourmodel-base & 86.5\% & -30.9\% & \underline{0.01\%} & \textbf{64.7\%} & \underline{26.5\%} & \textbf{35.7\%} \\
  - \ourmodel-warmup & \underline{33.7\%} & \textbf{21.9\%} & 49.1\% & 15.8\% & 37.5\% & 24.7\% \\
\hline
\end{tabular}
\label{tab:attack}
\end{table*}

\subsection{Evaluate VQA Models on Generated Distractors (\textit{RQ1})}
We answer RQ1: ``How does \ourmodel perform in comparison to other methods in terms of fooling pre-trained VQA models'' in this subsection. Table \ref{tab:attack} shows the performance degradation of pre-trained VQA models when presenting to distractors generated by different DG-VQA models. Since the three pre-trained VQA models use different architectures, the distractor generation model requires high generalization capability to confuse all three of them. As can be seen, all baseline models yield poor quality distractors in terms of \dacc. More specifically, \textit{Q-type prior} fails to fool any VQA models. \textit{Adversarial Matching} and \textit{LSTM Q+I} lack generalization capability, which is only able to abate one or two VQA models' accuracy in small margins. In contrast, our proposed \ourmodel methods yield significant improvements on all three pre-trained VQA models. It is worth noting that \ourmodel-base performs better on RevisitedVQA and MCB than \ourmodel-warmup, while worse on TellingVQA($\Delta \text{Acc} =-30.9\%$), It indicates that without the warmup process (\textit{i.e.} fine-tuning on correct answers), the agent model is vulnerable to overfitting (\textit{e.g.} $\text{Acc} =0.01\%$ for \ourmodel-base when it receives rewards from RevisitedVQA and tries to fool it at the same time). 

Furthermore, the larger \dacc of \ourmodel shows pre-trained VQA models provide an alternative knowledge source for DG-VQA, thus leading to \ourmodel capable of generating high-quality distractors without any training samples. However, only receiving rewards from one specific environment is not robust. We address this issue by incorporating the warm-up process, in which case it provides a smoother beginning probability distribution over the candidate pool. Therefore, it prevents the agent from falling into the biased local minima trap.

\begin{figure}[htbp!]
\centering
\begin{scaletikzpicturetowidth}{0.8\linewidth}
\begin{tikzpicture}[scale=\tikzscale]
\begin{axis}[
    ybar,
    enlargelimits=0.15,
    %legend style={at={(0.5,-0.2)},
    %  anchor=north,legend columns=-1,
    %  font=\tiny},
    legend style={at={(0.5,0.97)},
                anchor=north,legend columns=2,
                fill=none},
    ylabel={Accuracy},
    xlabel={Training dataset using different distractors},
    symbolic x coords={[O], [A], 0.5[O]+0.5[A]},
    yticklabel={\pgfmathprintnumber\tick\%},
    xtick=data,
    ymax=90.0, 
    nodes near coords,
    every node near coord/.append style={font=\small},
    nodes near coords align={vertical},
    every axis plot/.append style={fill},
    cycle list/Set2,
    ]
\addplot coordinates {([O],62.2) ([A],43.9) (0.5[O]+0.5[A],61.1)};
\addplot coordinates {([O],37.5) ([A],79.7) (0.5[O]+0.5[A],65.4)};
\addplot coordinates {([O],64.8) ([A],39.8) (0.5[O]+0.5[A],55.0)};
\addplot coordinates {([O],49.1) ([A],78.8) (0.5[O]+0.5[A],75.2)};
\legend{MCB@[O], MCB@[A], RevistedVQA@[O], RevistedVQA@[A]}
\end{axis}
\end{tikzpicture}
\end{scaletikzpicturetowidth}
\vspace{-0.25cm}
\caption{Data Augmentation Results}
\label{fig:augment}
\end{figure}

\subsection{Augmenting VQA model with Generated Distractors (\textit{RQ2})}

To answer RQ2, we utilize generated distractors as the augmented data to train more robust VQA models. In particular, we keep the correct answer of each input question image pair and swap the original distractors to the generated ones. MCB and RevisitedVQA are better suited for this setting since they take both correct and incorrect choices into consideration while training, while TellingVQA only takes the correct answer as input. Hence, we re-train MCB and RevisitedVQA models from scratch under two settings: 1) generated distractors alone; 2) the mixup of original and generated distractors. As a control group, we adopt the VQA models trained on the original distractors alone. All distractors are produced by our \ourmodel-warmup, which has been shown the most effective DG-VQA model. 

Figure \ref{fig:augment} reports the results, where the x-axis denotes on which training set the models are trained, and the y-axis denotes the model performance in terms of accuracy. $[O]$ and $[A]$ refer to the original data and the augmented data respectively. And $0.5[O]+0.5[A]$ denotes $50\%$ of all questions' incorrect alternatives are replaced by the generated distractors. Different bars indicate the VQA accuracy that models can achieve on a specific testing set, \textit{e.g.} the green bar denotes MCB model testing on the original distractors, while the orange bar denotes MCB models testing on the generated distractors. At first glance, we find that data augmentation training improves the models' performance on generated distractors. However, it hurts the model performance on the original test data. We observe a similar pattern for models trained solely on original distractors. As a balance, models trained on the union of augmented and original data achieve the best performance with the minimum $Acc@[O]$ drop of $1.1\%$ and the highest $Acc@[A]$ improvement by $27.9\%$. These results demonstrate the effectiveness of DG-VQA for training more robust VQA models.

\begin{table*}[htb]
\footnotesize \centering
\caption{Excerpts from sampled original and adversarial generated distractor choices. Green choices are correct answers. Bold texts indicate options chosen by the pre-trained RevisitedVQA model in Visual7W.}
\begin{tabularx}{\textwidth}{X X X X X}
\includegraphics[ width=0.8\linewidth, height=0.8\linewidth ]{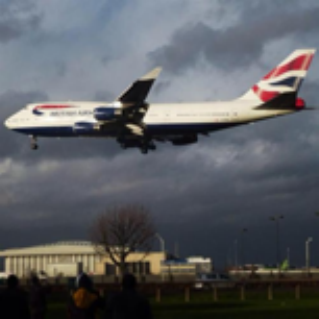}&
  \includegraphics[ width=0.8\linewidth, height=0.8\linewidth ]{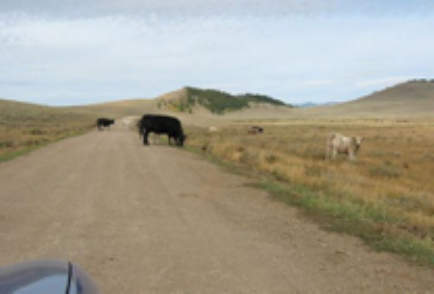}&
  \includegraphics[ width=0.8\linewidth, height=0.8\linewidth, keepaspectratio]{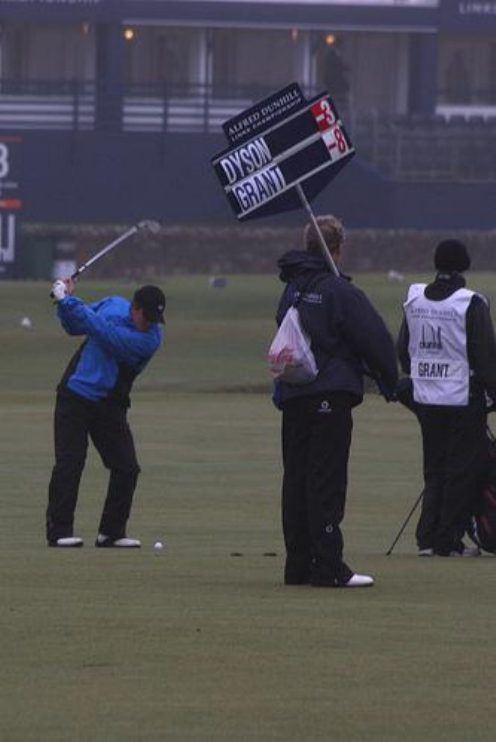}&
  \includegraphics[ width=0.8\linewidth, height=0.8\linewidth, keepaspectratio]{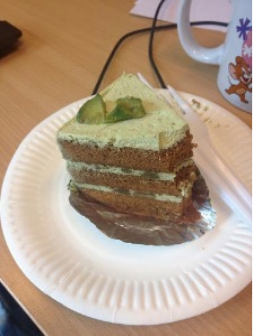}&
  \includegraphics[ width=0.8\linewidth, height=0.8\linewidth ]{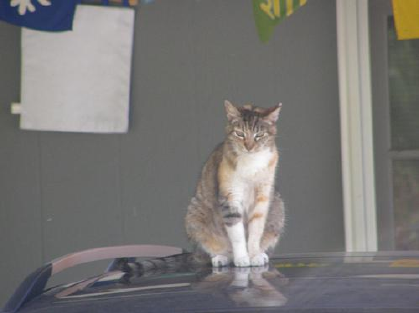} \\
  Q:What does the sky look like?& 
  Q:How many black cows are there?&
  Q:What sport are they playing?&
  Q:Why is there a piece missing?&
  Q:What two colors are in the flag directly above the cats head?\\
\hline
Original Choices\\
\textcolor{green}{A: \textbf{Stormy}} \greencmark& \textcolor{green}{A: \textbf{3}} \greencmark& \textcolor{green}{A: Golf}& \textcolor{green}{A: Someone ate some}& \textcolor{green}{A: Green and yellow}\\
A: Hazy& A: 9& A: \textbf{Baseball} \redxmark& A: It was removed& A: \textbf{Blue and white} \redxmark\\
A: Windy& A: 8& A: Hockey& A: \textbf{It was put somewhere else} \redxmark& A: Black and red\\
A: Sunny& A: 7& A: Basketball& A: Someone took it& A: Green and black\\
\hline
\multicolumn{4}{l}{Distractors by \textit{Adversarial Matching}}\\
A: Stormy& A: 3& A: \textbf{Golf} \greencmark& A: Someone ate some& A: Green and yellow\\
A: \textbf{Sky} \redxmark& A: Zero& A: Volleyball& A: Wood& A: Blue and black\\
A: Blue& A: 5& A: Playing soccer& A: \textbf{Glass} \redxmark& A: Blue and red\\
A: Cloudy& A: \textbf{0} \redxmark& A: Soccer& A: To rest& A: \textbf{Blue and white} \redxmark\\
\hline
\multicolumn{4}{l}{Distractors by \ourmodel-base}\\
A: Stormy& A: 3& A: Golf& A: Someone ate some& A: Green and yellow\\
A: Shadows& A: \textbf{Shadows} \redxmark& A: Shadows& A: Shadows& A: \textbf{Shadows} \redxmark\\
A: Daylight& A: During daylight& A: \textbf{During daylight} \redxmark& A: \textbf{Daylight} \redxmark& A: Daylight\\
A: \textbf{Shadow} \redxmark& A: In the daytime& A: Daylight& A: During daylight& A: In the daytime\\
\hline
\multicolumn{4}{l}{Distractors by \ourmodel-warmup}\\
A: Stormy& A: 3& A: Golf& A: Someone ate some& A: \textbf{Green and yellow} \greencmark\\
A: \textbf{Cloudy} \redxmark& A: Two& A: \textbf{Baseball} \redxmark& A: \textbf{To eat} \redxmark& A: Blue\\
A: Blue& A: Four& A: Soccer& A: To cook& A: Legs\\
A: Clouds& A: \textbf{One} \redxmark& A: Tennis&A: For display& A: Orange\\
\hline
\end{tabularx}
\label{tab:case_study}
\end{table*}

\subsection{Case Study (\textit{RQ3})}\label{subsec:casestudy}

The case study is critical to answering \textit{RQ3}, since the pre-trained VQA models' performance degradation and the data augmentation effectiveness only indirectly validate the quality. Meanwhile, the widely used text generation measurements such as BLEU, ROUGE are not applicable either, because high quality is not necessarily related to n-gram similarity between original distractors and generate distractors. Thus, We collect textual distractor choices generated by baselines and the proposed methods, as can be seen in Table~\ref{tab:case_study}. We further analyze them in-depth, and have observed the following factors lead to high-quality and challenging distractors generated by \ourmodel:

\noindent \textbf{Concept Similarity}: It is not surprising that \ourmodel learns the strategy to replace correct answers with conceptually similar terms, as humans follow the same strategy to come up with distracting choices. As we can see, distractors generated by \ourmodel and the correct answers almost belong to the same concept categories. For example, ``baseball'', ``soccer'', ``tennis'' (distractors by \ourmodel-warmup of the third example) and ``golf'' (correct answer of the third example) are sport terms. And all distractors produced by \ourmodel-warmup for the second question ``how many black cows are there'' are all numbers, which belong to the same category of the correct answer: ``3''.

\noindent \textbf{Context Matters}: Another critical factor is input context. In the first column of Table \ref{tab:case_study}, both distractors of the original dataset and augmented ones are adjectives to describe the weather. However, ``cloudy'' is better than ``stormy'' to depict the picture, compared to ``hazy'', ``windy'' and ``sunny''. Under the original choice setting, the defender can select the correct answer. But once encountered with the generated distractors, it is confusing and misleadingly pick ``cloudy'' as the answer. Tackling vision and language tasks needs multimodal cognitive ability. In the DG-VQA task, a system should comprehensively utilize information from both the given questions and the images.

\noindent \textbf{Attack the Weaknesses and Improve}: Our architecture is able to receive feedback from the defender (pre-trained VQA model). It is common to exploit opponents' weaknesses to defeat them. By analyzing judgment scores of the alternatives, the distractor generator identifies the differences between the hard and the easy ones. Examples of this can be found in distractors generated by \ourmodel-base (see the first question in Table \ref{tab:case_study}). It seems that our system generates easy-to-human distractors like ``shadows'' or ``daylight''. However, the defender is observed to be confused by them. A similar phenomenon has also been observed in \cite{43405}, where neural networks are vulnerable to small perturbations on input images while humans can easily distinguish them. Our model is able to identify such tricky weaknesses of defender models and exploit them. Moreover, by considering these weaknesses for the next round of training, a model's robustness is improved. 

In summary, the case study supports that our method in fact outputs high-quality distractors by considering all together with the semantics of the correct answer, the information of the context, and the feedback from the trained VQA models as an alternative knowledge source.

\section{Conclusion}
\label{sec:conclusion}

In this work, we introduce the novel DG-VQA task. These generated ``hard negative" distractors are significant since deep networks have been applied in many real-life and safety-sensitive environments. One major challenge for DG-VQA is the sparsity of training samples. To address it, we developed the policy gradient-based \ourmodel, where pre-trained VQA models serve as the alternative knowledge source to guide the distractor generation. Furthermore, the generated distractors can provide insights into factors that cause VQA models vulnerable.

Recent advances in text and image retrieval have enabled many multi-modality applications to deal with open-world, knowledge-based scenarios. Instead of relying on the established knowledge base, \ourmodel paves a new pathway for future research that leverages pre-trained VQA models as an underlying multi-modal knowledge base. Whilst learning pre-training tasks, these models may also be storing latent cross-modality knowledge present in the training data. Compared to established structured KBs, pre-trained models have many advantages, such as they do not require a pre-defined schema or ad-hoc canonicalization process, thus enabling them easy to extend to different domains.

% We hold the view that the DG-VQA task and the adversarial training towards distractor generation for visual questions pave a new pathway for further research in anti-adversarial and explainable VQA. There are several caveats of our method that is worth mentioning. For instance, the alternatives generated are less diverse and the lingering concern of over-fitting by our proposed MPLR and MPLR+Pre-train methods. It sparks future directions such as distractor generation for VQA with explicit and explainable reasons. 

{\small
\noindent \textbf{Acknowledgement.}
This work was supported by the National Science Foundation under Grant CNS-2101052, IIS-2132724 and IIS-1750082. 
}

%%%%%%%%% REFERENCES
{\small
\bibliographystyle{ieee_fullname}
\bibliography{egbib}
}

\end{document}